\documentclass[conference]{IEEEtran}
\IEEEoverridecommandlockouts
\usepackage{verbatim}
\usepackage{url}
\usepackage{hyperref}
\usepackage{tabularx}
\usepackage{float}
\usepackage{cite}
\usepackage{amsmath,amssymb,amsfonts}
\usepackage{algorithmic}
\usepackage{graphicx}
\usepackage{textcomp}
\usepackage{xcolor}
\usepackage[numbers]{natbib}
\usepackage{adjustbox} 
\usepackage{tikz}
\usetikzlibrary{shapes.geometric, arrows.meta, positioning, shadows, calc}
\usepackage{graphicx} 

\usepackage[labelfont=bf,textfont=md]{caption}
\usepackage{authblk}

\def\BibTeX{{\rm B\kern-.05em{\sc i\kern-.025em b}\kern-.08em
    T\kern-.1667em\lower.7ex\hbox{E}\kern-.125emX}}
\begin{document}

\title{Dynamic Activation with Knowledge Distillation for Energy-Efficient Spiking NN Ensembles\\
}
\author[3$^{\ast}$]{Orestis Konstantaropoulos}
\author[3$^{\ast}$]{Theodoris Mallios}
\author[1,2,3]{Maria Papadopouli}

\affil[1]{Department of Computer Science, University of Crete, Heraklion, Greece}
\affil[2]{Institute of Computer Science, Foundation for Research \& Technology-Hellas, Heraklion, Greece}
\affil[3]{Archimedes, Athena Research Center, Athens, Greece}

\maketitle

\begingroup
\renewcommand{\thefootnote}{\fnsymbol{footnote}}
\footnotetext[1]{Alphabetical ordering; Equal contribution. Contact author: Maria Papadopouli (maria@csd.uoc.gr).}
\endgroup

\begin{abstract}
While foundation AI models excel at tasks like classification and decision-making, their high energy consumption makes them unsuitable for energy-constrained applications. Inspired by the brain's efficiency, spiking neural networks (SNNs) have emerged as a viable alternative due to their event-driven nature and compatibility with neuromorphic chips. This work introduces a novel system that combines \textit{knowledge distillation} and \textit{ensemble learning} to bridge the performance gap between artificial neural networks (ANNs) and SNNs. A foundation AI model acts as a teacher network, guiding smaller \textit{student SNNs} organized into an ensemble, called Spiking Neural Ensemble (SNE). 
SNE enables the disentanglement of the teacher's knowledge, allowing each student to specialize in predicting a distinct aspect of it, while processing the same input. 
The core innovation of SNE is the \textit{adaptive }activation of a subset of SNN models of an ensemble, leveraging knowledge-distillation, enhanced with an informed-partitioning (disentanglement) of the teacher's feature space. 
By \textit{dynamically activating only a subset }of these student SNNs, the system balances accuracy and energy efficiency, achieving substantial energy savings with minimal accuracy loss. Moreover, SNE is significantly more efficient than the teacher network, reducing computational requirements by up to \textit{20x} with only a 2\% drop in accuracy on the CIFAR-10 dataset.
This disentanglement procedure achieves an accuracy improvement of up to 2.4\%
on the CIFAR-10 dataset compared to other partitioning schemes.
Finally, we comparatively analyze SNE performance under noisy conditions,
demonstrating enhanced robustness compared to its ANN teacher. In summary, SNE offers a promising new direction for energy-constrained applications.

\end{abstract}


\section{Introduction}
Foundation AI is repeatedly breaking ground in computer vision and machine learning \cite{b30}, \cite{b31}, with advancements at dramatic speed across various domains, including image and video classification, semantic segmentation, depth estimation, image captioning, and decision-making. However, training and deploying foundation AI models demands extraordinary amounts of energy and data \cite{b32}. Even with efforts to reduce model sizes, inference still requires hundreds of Watts, making these models impractical for deployment at the edge, where energy efficiency is crucial. On the other hand, the human brain is remarkably energy-efficient, consuming about 20 watts of power, impressive given its computational capabilities, involving billions of neurons firing in complex patterns to process sensory input, control movement, and enable cognition. Spiking neural networks (SNNs), inspired by biological neuronal networks, provide a promising alternative for achieving energy-efficient intelligence \cite{b33}. They use binary spiking signals (0 for no activity and 1 for a spiking event) to facilitate communication between neurons. SNNs can operate efficiently on neuromorphic chips by performing spike-based accumulate (AC) operations, eliminating the need to process zero values in inputs or activations (i.e., they are event-driven). This enables SNNs to consume significantly less power compared to artificial neural networks (ANNs), which rely on energy-intensive multiply-and-accumulate (MAC) operations typically performed on dense computing hardware like GPUs. With the advent of neuromorphic chips \cite{b34}
the integration of neuromorphic processors into everyday devices is becoming increasingly plausible.

In our research, we envision a system that harnesses the transferability and generalization capabilities of state-of-the-art AI foundation models to enhance neuromorphic architectures that combine energy efficiency with high accuracy. To achieve this, we adopt a knowledge distillation approach, where the knowledge from foundation AI models is transferred to neuromorphic architectures. In this framework, large-scale foundation AI architectures serve as \textit{teachers} that guide \textit{smaller}, neuromorphic, or more broadly neuroscience-inspired, student architectures enabling them to emulate teacher's performance. Specifically, we propose an innovative neuromorphic architecture trained using \textit{knowledge-distillation}: 
a powerful ANN acts as a single-teacher model distilling its knowledge to an \textit{ensemble of small student SNNs}, called Spiking Neural Ensemble  (\textbf{SNE}). This combination of knowledge distillation and ensemble learning can significantly improve the energy efficiency and accuracy of SNNs, aiming to reduce the performance gap between ANNs and SNNs.
Our approach leverages the Single-Teacher, Multiple-Student paradigm, where each student learns to mimic a \textit{distinct subset} of features from the teacher network. 
Specifically, the teacher’s feature space—the final layer just before the classification head—is partitioned into distinct subsets, with each student responsible for replicating one of them.

Each student processes the \textit{same input image}, generating a feature vector. 
These feature vectors are concatenated to form the ensemble’s full feature representation. During training, this concatenated vector is compared against the teacher’s full feature vector to ensure alignment with the teacher’s performance and then it is passed through a classification head. In testing, the ensemble utilizes its \textit{collective feature} representation for classification. 

\textbf{Innovation}  To the best of our knowledge, it is the first study of knowledge-distillation on an \textit{\textbf{ensemble}} of \textbf{SNN} students. 
We propose a set of techniques that improve the single-teacher, multiple-student knowledge-distillation paradigm. Specifically, we introduce a novel \textbf{fine-tuning procedure for the teacher} to \textit{disentangle its feature space}, ensuring that each student learns a \textit{distinct part of the teacher}’s information.
Inspired by the brain’s sparsity and modularity, SNE is enabled to \textit{dynamically} determine the ensemble's size based on the accuracy-energy trade-off. Since each student is assigned a distinct sub-task, \textit{not all} student SNNs need to be active to accomplish a task effectively. By organizing the students into an ensemble, SNE can \textit{dynamically} select the subset of students to activate, thereby improving energy efficiency.

We demonstrate that our method achieves remarkable energy efficiency without compromising performance. For example, SNE reduces computational requirements from 398M FLOPS in the teacher network to just 18.4M FLOPS (a \textit{20x reduction}) with only a \textit{2\% drop in accuracy} on the CIFAR-10 dataset. 
Additionally, by selecting the appropriate number of active students in the student ensemble, SNE can achieve up to a 65\% decrease in energy consumption with a mere 2.07\% drop in accuracy. 
In case of low-noise input, the ensemble demonstrates consistent performance and exhibits enhanced robustness in high-noise conditions.
Furthermore, the proposed disentanglement procedure achieves an accuracy improvement of up to 2.4\%
on the CIFAR-10 dataset compared to other partitioning schemes.
The remainder of this paper is structured as follows:
Section \ref{sec:back} reviews the main paradigms and approaches in training efficient SNNs and knowledge distillation from (teacher) ANNs to SNNs students. Section \ref{sec:system} presents the proposed architecture while Section \ref{sec:perf} evaluates its performance. Finally, Section \ref{sec:disc} summarizes the main contributions and future work plans.

\section{Background}\label{sec:back}
\subsection{Training Efficient Spiking Neural Networks}\label{sec:snn}
Training SNNs is challenging due to the non-differentiable nature of spikes. Current mainstream approaches to SNN training can be classified into two broad categories, namely\textit{ ANN-to-SNN conversion} and \textit{direct SNN training}, described below. A common approach for training SNNs is to leverage the well-established techniques used in ANNs by converting a high-performing ANN into an equivalent SNN with similar accuracy \cite{b1, b2, b3, b4, b5, b14, b15}. This process typically involves replacing the ANN's ReLU activation functions with spiking neurons and fine-tuning the SNN to approximate the ANN's outputs. Although this approach can achieve high accuracy, it often requires a large number of time steps for the SNN to replicate the ANN’s outputs effectively \cite{b2, b4, b5}. This reliance on extended simulation time limits its applicability in energy-constrained environments or latency-critical applications, such as real-time inference. The direct SNN training using surrogate gradients (SG) \cite{b7, b8, b16, b17, b20, b23, b24} addresses the non-differentiability of spikes by approximating the gradient of the spiking function with a smooth surrogate function during the backward pass. Combined with backpropagation through time (BPTT), it enables weight updates across multiple time steps. While direct training provides more freedom in the design of SNNs, it often leads to worse performance compared to ANNs, due to the intrinsic complexity of spiking dynamics and limitations of surrogate approximations.

\subsection{Knowledge Distillation from ANN Teachers to SNN Students} \label{sec:kd-ANN-SNN}
Although knowledge distillation (KD), a powerful framework for transferring the knowledge of a large, complex network (teacher) to a smaller, more efficient network (student) \cite{b6},
was originally developed for ANN-to-ANN compression, variants have recently been proposed for SNNs, demonstrating improved SNN training \cite{b11, b12, b13}. 

\textit{Feature-based} distillation matches the \textit{intermediate representations} of the teacher with the \textit{intermediate representations} of the student. In contrast, \textit{output-based }distillation aligns the final output of the two models. The alignment is usually performed by incorporating a distance term in the loss function which is minimized. These methods have demonstrated that knowledge transfer is possible, despite the differences between ANNs and SNNs, and can greatly improve SNN performance. More specifically, by using feature distillation in every single layer of an SNN student network and aligning them with representations of an ANN teacher, the SNN student achieved 93\% accuracy on CIFAR-10 with only 4 timesteps per example, setting a new state-of-the-art for spiking convolutional models \cite{b12}.
\subsection{Ensemble Learning in SNNs}\label{sec:ensemble}
Ensemble-training methods, typically used in ANNs, have also been applied to SNNs to enhance performance and robustness \cite{b9, b10}.
Most methods aggregate the predictions of multiple models with the same architecture but different initial weights and leverage a majority voting algorithm to make the final prediction and improve overall accuracy. In the context of SNNs, ensemble techniques are often used to increase the accuracy of a model \textit{without} taking into consideration the increased computational costs of training its multiple instances \cite{b19}. In this work, we aim to develop models that achieve comparable accuracy and computational cost to single-model architectures, while providing the \textit{added flexibility to balance accuracy and energy consumption}.

In traditional ANNs, ensemble techniques, like the Teacher-Class Network (TCN) \cite{b18}, where a set of smaller feature-specific networks are trained under the guidance of a teacher model, have been proposed. Such techniques offer modularity, enabling the use of the necessary modules, without degrading the performance compared to a single-student architecture. 

We aim to adapt this technique to SNNs, where multiple spiking models are trained in conjunction with a teacher network to optimize performance. This combination of knowledge distillation and ensemble learning can significantly improve the energy efficiency and accuracy of SNNs, aiming to reduce the gap between ANN and SNN performance. 

\section{Single-Teacher Multiple-Student SNN Ensemble (SNE)}\label{sec:system}
We introduce a novel training framework that combines recent advancements in Spiking Neural Network (SNN) training with the knowledge distillation paradigm, employing a single-teacher model and multiple student models. The teacher model is a standard deep convolutional neural network (CNN), while the student models are SNNs, with shallower architectures than the teacher, trained to mimic the feature representations generated by the teacher. Each student learns a distinct subset of the features produced by the teacher's final layer, just before the classification head. The partitioning of the feature set, and thus its allocation to students, could be performed using different techniques, as discussed in Section \ref{sec:disentangle}. 

The features extracted by the student SNNs are then concatenated and fed into a classification head for downstream tasks. This ensemble-based approach achieves comparable performance to a single-student SNN with the same total computational cost. Furthermore, the method offers flexibility in balancing computational cost and model performance by selecting the number and depth of the used student models.
\begin{figure}[t]
    \adjustbox{width=\columnwidth, height=6cm, keepaspectratio}{
        \input{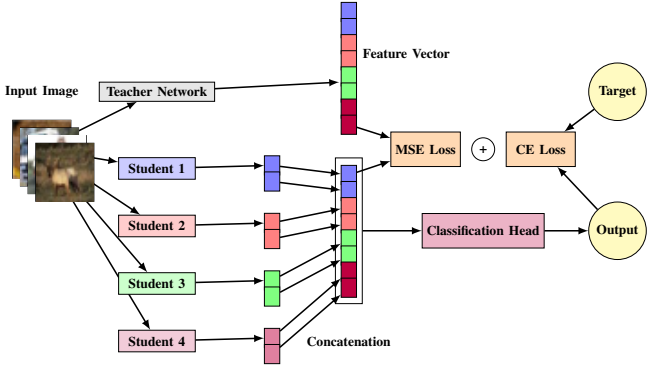}
    }
    \vspace{-\baselineskip}
    \caption{\textbf{Training a Student Ensemble.} The input image is fed into the teacher network and into each student network. Each network computes a vector of features. The features of the students are concatenated and the Mean Square Error (MSE) is computed between the teacher's feature vector and the concatenated feature vector (obtained from the student ensemble). The concatenated feature vector of the ensemble is fed into a Linear classification head which produces an output vector that is used to calculate the Cross Entropy Loss from the ground truth vector for the example.}
    \label{fig:ensemble}
\end{figure}
\subsection{Knowledge Distillation for SNN Models}
The neurons of the student SNNs follow the Leaky Integrate and Fire (LIF) model \cite{Mihalas2009:LIF,b21}. Each neuron has a state divided into three processes, namely, charging, discharging, and resetting. The charging process can be expressed as in Eq. \ref{eq:1}, where \(V[t]\) denotes the membrane potential of the neuron model at time $t$, \(\tau_m\)  is the membrane time constant, while \(X[t]\) indicates the external input current, which corresponds to the weighted sum of the spikes fired by the neurons in the previous layer. Here, \(H[t]\) represents the membrane potential after charging but before firing. The process of firing binary spikes \(S[t] \) in the neuron model is represented by a Heaviside step function \(\Theta\) (the value of which is zero for negative arguments and one for positive arguments), as described in Eq. \ref{eq:2}, where \(V_{th}\) represents the fired threshold of the membrane potential. The discharging process is expressed in Eq. \ref{eq:3}, where \(V_{reset}\) represents the value to which the membrane potential resets after firing.
\begin{figure}[t]
    \centering
    \adjustbox{width=\columnwidth, height=6cm, keepaspectratio}{
        \includegraphics{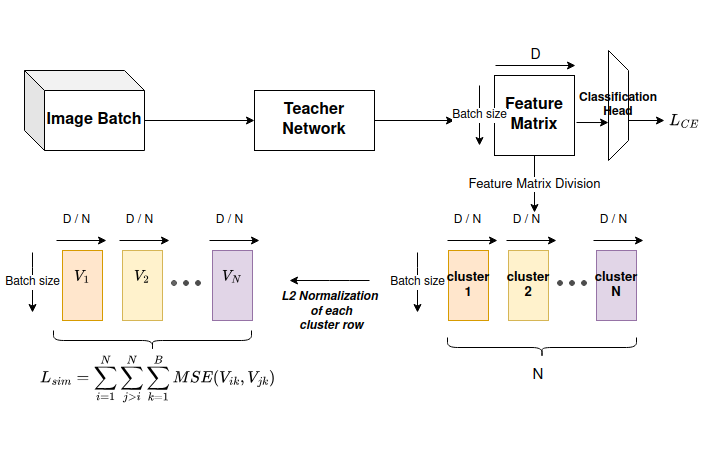}
    }
    \vspace{-10mm} 
    \caption{\textbf{Fine-tuning of the Teacher Network}. The teacher network is fine-tuned to naturally partition its feature space: in an online iterative manner, for each batch, the feature matrix is divided into $N$ clusters: a cluster includes a number of feature columns. After normalizing each cluster row, to increase the separability between clusters, we employ a loss metric based on the mean distances of each feature row of a cluster from the corresponding feature rows of all other clusters \(L_{sim}\). The loss function in the fine-tuning of the teacher's network combines the loss for the primary classification task \(L_{CE}\) and a negatively weighted \(L_{sim}\), promoting better feature clustering while maintaining classification accuracy.}
    \label{fig:teacher_ft}
\end{figure}

\begin{equation}
    H[t] = V[t-1] + \frac{1}{\tau_m}(X[t] -V[t-1])
    \label{eq:1}
\end{equation}

\begin{equation}
    S[t] = \Theta(H[t] - V_{th})
    \label{eq:2}
\end{equation}

\begin{equation}
    V[t] = H[t] \cdot (1 - S[t]) + V_{reset} \cdot S[t]
    \label{eq:3}
\end{equation} 

To train SNNs using gradient descent algorithms similar to those in ANNs, we must address the non-differentiable nature of the Heaviside step function \(\Theta\) used in the neuronal model of SNNs. During the backward pass of backpropagation (BP), we employ a surrogate gradient approach, approximating \(\Theta(x)\) with a similar yet differentiable function \(\sigma(x)\), such as a sigmoid function. In BP, the undefined gradient of \(\Theta(x)\) is replaced by the gradient of \(\sigma(x)\), allowing effective weight updates.

The student SNNs are trained using a dual-loss function \(L\) to optimize their performance.
\begin{equation}
    L = L_{CE} + \alpha * L_{KD}
    \label{eq:4}
\end{equation}
The first loss term \(L_{CE}\) is the standard cross-entropy loss, which directly addresses the primary classification task, assessing the difference between the output of the model and the ground truth label, so that the model learns to classify the correct class. 

The second loss term \(L_{KD}\) inspired by \cite{b12} leverages the knowledge distillation paradigm, where the ensemble aims to approximate the feature representations extracted from the teacher model. Specifically, for each example, the feature extraction from the teacher is performed by conducting a forward pass of the example through the teacher model up to its designated feature layer. The distillation task is implemented using a mean squared error (MSE) loss between the teacher’s feature vector and the ensemble's feature vector, encouraging the ensemble to regress toward the teacher's feature space (e.g., Eq. \ref{eq:5}).
\begin{equation}
    L_{KD} = \sum_{i=1}^{D} (v_i - k_i)^2,
    \label{eq:5}
\end{equation}
where \(D\) is the number of dimensions of the feature vector, \(v\) is the feature vector of the teacher network, and \(k\) is the feature vector of the student network. The hyperparameter \(\alpha\) controls the impact of the teacher on the student; The special case of 
\(\alpha\) = 0 corresponds to training a traditional SNN without knowledge distillation. The specific loss  \(L_{KD}\) used in our case is discussed in Section \ref{sec:stud_ens}.

To utilize the teacher's knowledge and have increased performance when compared to traditional SNN training, we observed that high \(\alpha\) values are required so that the \(L_{CE}\) term does not dominate the total loss and the students regress towards the teacher's feature space and then learn to classify. The results presented here are based on $\alpha=2$.

\subsection{Student Ensemble}\label{sec:stud_ens}
Rather than distilling knowledge from the teacher model into a single-student network, our proposed method employs an \textit{ensemble} of student models. Each student is assigned the task of learning a specific \textit{subset} of the teacher's feature vector, which is partitioned into \textit{N }distinct (\textit{non-overlapping}) sub-vectors, where \textit{N} corresponds to the number of students. For instance, in the case of two students and a feature vector of size 100, the first student processes features corresponding to indexes [1, 50], while the second processes indexes [51, 100]. 

The teacher's feature space \(S = \{1,..,D\} \) is partitioned into \textit{N} distinct subsets \(S_i\) and each subset is assigned to a different student, in an ensemble of \textit{N} students. We will name the partitioning of random equal-sized subsets \textit{\textbf{fixed}} (with \textit{no} feature disentanglement).

During the inference phase, the teacher generates feature vector \(\bold{v}\) and each student generates a predicted feature sub-vector \( s_i\). The Mean Square Error (MSE) is calculated between these predicted sub-vectors and their corresponding teacher sub-vectors \(v_i = \{\bold{v}_k : k \in S_i\} \). The outputs of all students are subsequently concatenated to form a comprehensive representation, which is then input into a linear classification head \(CH\) to produce the final prediction: $s = s_1 \oplus s_2 \oplus ... \oplus s_N$; ensemble's output: $e_o=CH(s)$.
%
The student ensemble is trained concurrently using the loss function defined in Eq. \ref{eq:4}, where now
\begin{equation}
    L_{KD} = \sum_{i=1}^N \sum_{k=1}^{|S_i|}(v_{ik}-s_{ik})^2 \label{eq:9}
\end{equation}
and \(L_{CE}\) is the cross-entropy loss assessing the difference between the \(e_{o}\) and the ground truth label. Backpropagation is performed using a Heaviside step function as a surrogate for spikes.

The features dimensions \(S_i\) assigned to each student are determined based on indexing. However, a more ``strategic" grouping of feature dimensions could improve students' ability to learn the assigned feature vector \( v_i \), as discussed below (in Section~\ref{sec:disentangle}).
\subsection{Feature Disentanglement}\label{sec:disentangle}
This section introduces one of the core ideas of this work, based on the assumption that higher-order representations of input data can form (ideally well-separated) \textit{clusters of features}, which will then be assigned to \textit{different} students. We demonstrate how the assignment of feature dimensions to students can be \textit{adaptive} and \textit{data-driven}. Rather than arbitrarily dividing the teacher's feature dimensions, we propose two techniques for feature partitioning aimed at improving student training, namely, the \textit{frozen} teacher and the \textit{fine-tuned} teacher.


In the frozen-teacher approach, we apply clustering algorithms in order to optimally divide the \textit{frozen} teacher's feature dimensions. Specifically, we generate a feature matrix by passing the entire training dataset through the teacher network. Clustering algorithms, such as K-means or agglomerative clustering, are then applied to the columns of this matrix based on the cosine similarity. Each cluster corresponds to a subset of feature dimensions (i.e., columns of the feature matrix), which will be assigned to student models. Agglomerative and k-means clustering may produce clusters of varying sizes, an imbalance that will lead to unequal workloads among students. To address load-balancing requirements, we can enforce \textit{equal-sized clusters} employing a variation of k-means clustering based on a greedy heuristic.


The second approach assumes that assigning correlated feature dimensions to students enhances their ability to learn. To this end, we propose fine-tuning the teacher network to refine the partitioning of its feature space, forming sub-features that: \textbf{i)} contain more correlated feature dimensions and \textbf{ii)} are more separable from other sub-features. To achieve this, we introduce an additional loss term during the fine-tuning process to ``guide" the teacher to produce features explicitly partitioned for the desired number of students \(N\). At each fine-tuning iteration, the feature space is evenly divided into \(N\) clusters \(V_i\). For example, cluster $i$ includes columns \(D/N*i\) to \(D/N*(i+1)\) of the feature matrix \textbf{\(V\)}. 


To ensure that the feature subspaces are distinct and well-separated, we apply a similarity loss. This loss penalizes the \(k\)-th feature row of the cluster \(i\) based on its proximity to the  \(k\)-th feature rows of all other clusters \(j\), \(i\neq j\) using the Euclidean distance metric:
\begin{equation}
L_{sim} = \sum_{i=1}^N \sum_{j > i}^N \sum_{k=1}^{B} MSE(V_{ik}, V_{jk}) \label{eq:10}
\end{equation}
where \(V_{ik}\) is the k-th row of cluster \(i\) and \(B\) is the size of each batch. The total loss of the teacher's fine-tuning procedure combines the classification loss \(L_{CE}\) and
the similarity loss as
\(Loss = L_{CE}+\lambda L_{sim}\), where \(\lambda\) is negative and controls the contribution of the similarity term. 

Figure \ref{fig:teacher_ft} illustrates the process, where \(N\) represents the number of students and \(D\) the dimensionality of the teacher's feature space.  After this fine-tuning process, SNE is trained as described in Section \ref{sec:stud_ens}.

\section{Performance Analysis}\label{sec:perf}
\begin{figure}[t] 
    \centering
    \includegraphics[width=\linewidth]{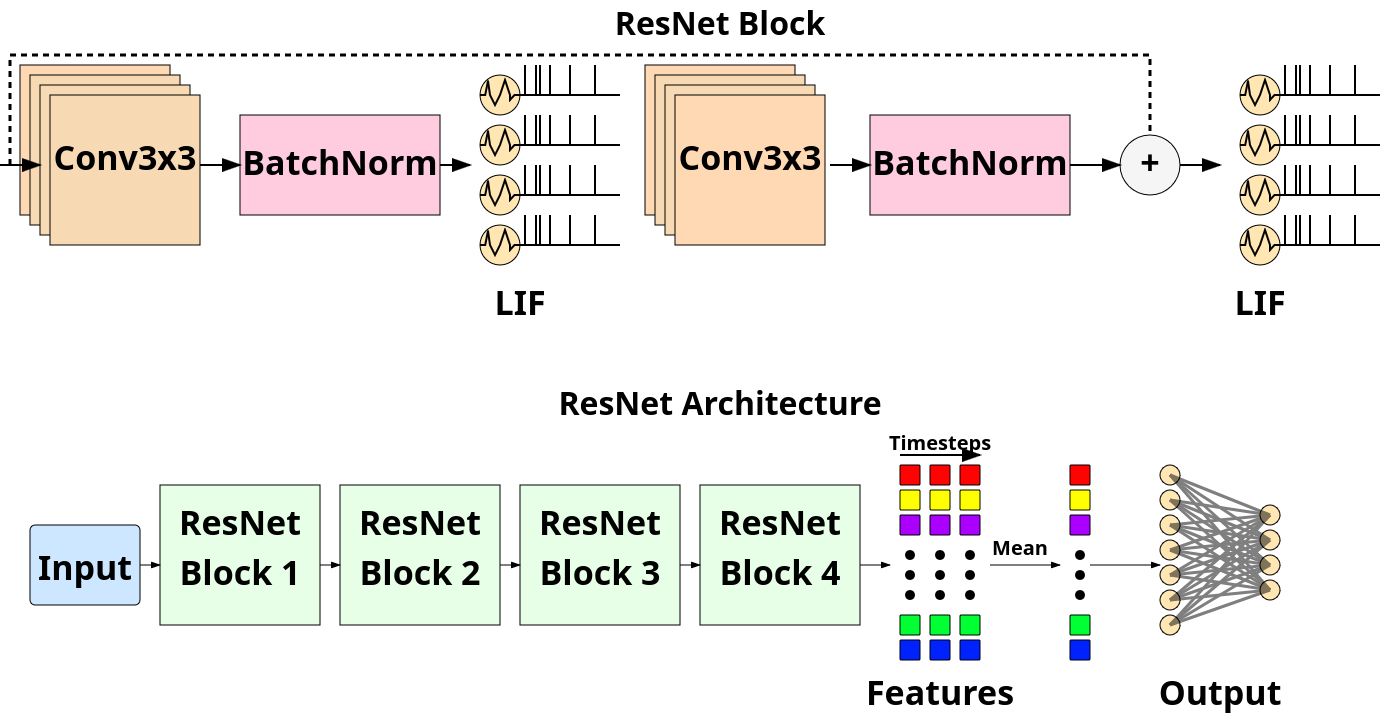}
    \caption{Overview of the Spiking ResNet architecture.}
    \label{fig:res_over}
\end{figure}
This section evaluates SNE using 
the CIFAR-10\footnote{CIFAR-10 consists of color images in 10 different classes. The dataset has 50,000 training images and 10,000 test images. The pixel values of images are scaled to be in the interval [0, 1].} dataset.
For the teacher architecture, we employ the ResNet and VGG architectures. All SNN student networks utilize Leaky Integrate and Fire nodes which serve as a replacement for the ReLU activation function in their corresponding ANN architecture. 
To quantify the computational demands of the different models, we measure two fundamentally different approaches to processing neural activations, namely the \textit{Multiply-Accumulate (MAC)} and \textit{Accumulate (AC)} operations during inference. MAC operations, prevalent in traditional ANNs, involve both multiplication and addition steps, where each input is multiplied by its corresponding weight before being accumulated into the running sum. In contrast, AC operations, used in SNNs, primarily rely on simple accumulation of incoming spikes without multiplication, making them computationally more efficient but potentially less expressive. Both metrics count the number of floating point operations (FLOPs) that are performed. The tracking of the MAC and AC operations is done with the help of the SyOPs \cite{b22} library.
Across our experiments, we utilize the PyTorch and the SpikingJelly \cite{b21} framework to evaluate our proposed methods.
The raw pixel values are used as input for the SNN models. We repeat the input for \(T = 4\) timesteps. After the processing of the SNNs, the firing rate of the final layer neurons, defined as their mean activation over $T$ timesteps, is used as input for the classification head.
\subsection{On the Number of Students}\label{sec:num_students}
We will first consider \textit{fixed} feature partitioning (\textit{no} feature disentanglement), and assess the performance of the SNE in terms of accuracy and number of FLOPs, for different numbers of students, compared to a teacher model, a single, two- or four-student SNN ensemble for both VGG and ResNet architectures (Tables \ref{tab:resnet_perf} \ref{tab:vgg_perf}).

To accommodate for a fair comparison, we develop an ensemble with the same computational cost as a single student architecture by forming students of the ensemble with reduced width, depth, or both. For an ensemble with only two students, we reduce the depth of the ResNet and VGG architectures from depths of 11 and 18 layers to 5 and 10, respectively.
Given that CNNs with two to four layers have degraded performance, we reduce the number of channels of the CNNs of ResNet10 and VGG5, forming the ResNet10mini and VGG5mini architectures, respectively, as follows:
Both ``mini" versions have the same depth as their corresponding (original) architectures. In addition, the last two CNN layers of VGG5mini are of half the width, while the number of channels of the first CNN layer of ResNet10mini is reduced from 64 channels to 54, and thus, its subsequent layers, which have the base width multiplied by an expansion index, are also reduced.

\begin{figure}[t]
    \centering
    \includegraphics[width=\linewidth]{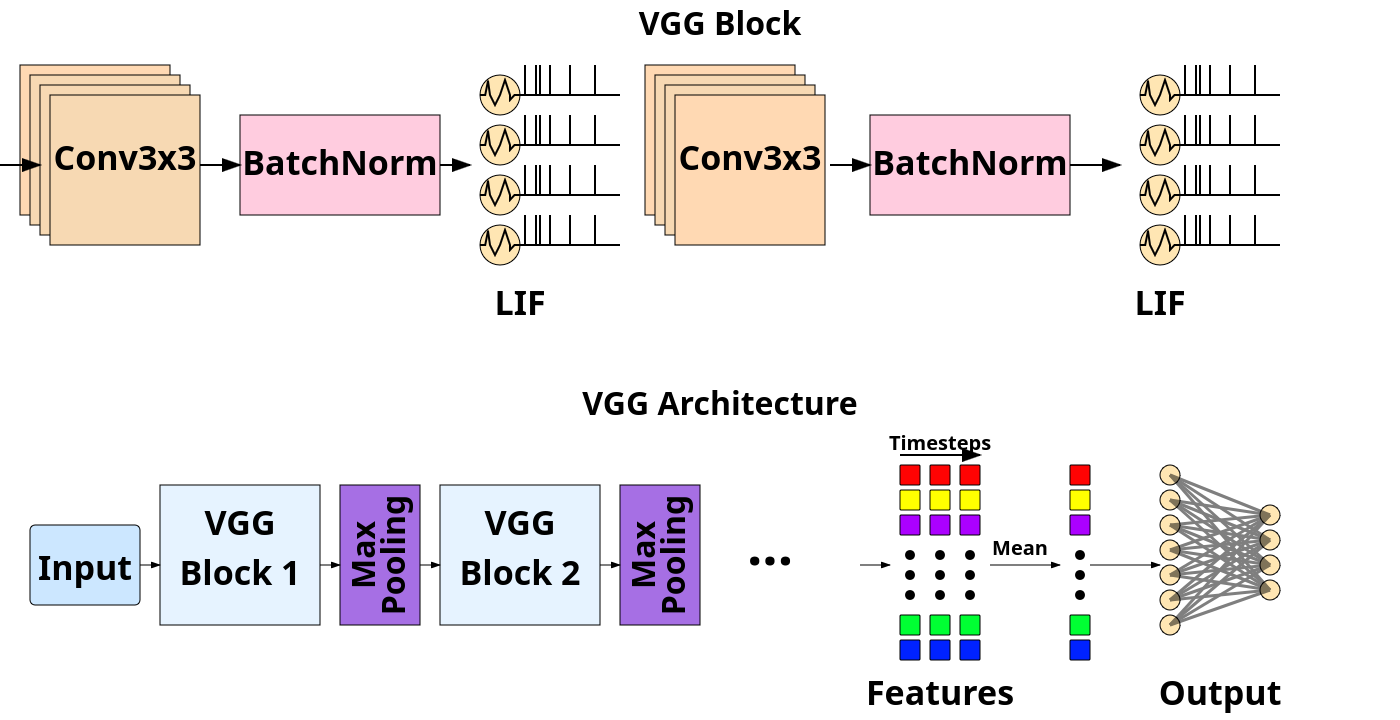}
    \caption{Overview of the Spiking VGG architecture.}
    \label{fig:vgg_over}
\end{figure}

Both architectures are built upon fundamental blocks that consist of convolutional layers followed by a batch normalization layer. However, instead of employing traditional activation functions, like ReLU, which are commonly used in ANNs, these blocks utilize Leaky Integrate-and-Fire (LIF) nodes. The LIF nodes generate spike trains for each feature, across all time steps. The primary distinctions between the two architectures lie in their structural design: ResNet incorporates skip connections to facilitate gradient flow and mitigate vanishing gradients, while VGG employs maxpooling layers between its blocks to down-sample feature maps. In VGG, the maxpooling layers process one timestep at a time, generating smaller-dimensional kernels for each convolutional channel as the data propagates through the network. 

The performances of the VGG and ResNet architectures with 2- and 4-student ensembles without disentanglement showcase a minor reduction in the accuracy compared to a single-student approach (Tables \ref{tab:vgg_perf} and \ref{tab:resnet_perf} and \ref{tab:vgg_perf}).
An ANN VGG19 teacher and an ANN ResNet18 teacher are employed for the SNN VGG and the ResNet architectures, respectively. It is evident that the use of Knowledge Distillation increases the performance of the Single Student by 0.5-1.0\%, consistent with \cite{b12}, and the two- and four-student ensembles can have comparable evaluation accuracy using fixed partitioning. 

\begin{table*}[t]
    \begin{minipage}[t]{\columnwidth}
        \centering
        \resizebox{\columnwidth}{!}{
        \begin{tabular}{|c|c|c|c|c|}
            \hline
            \textbf{Type} & \textbf{Architecture} & \textbf{Accuracy} & \textbf{Parameters} & \textbf{FLOPS} \\ \hline
            ANN & ResNet18       & 94.27\%  & 11.3M  & 555M MACs \\ \hline
            SNN & 1xResNet18 (\textit{without} KD) & 92.85\% & 11.3M & 165M ACs \\ \hline
            SNN & 1xResNet18       & 93.86\%  & 11.3M  & 165M ACs \\ \hline
            SNN & 2xResNet10     & 93.03\%  & 9.9M   & 107M ACs \\ \hline
            SNN & 4xResNet10mini & 93.22\%  & 11.9M  & 154M ACs \\ \hline
        \end{tabular}
        }
        \vspace{0.2cm}
        \caption{Validation accuracy, total parameter count and estimated FLOPs for a single forward pass of ResNet-based architectures, comparing ANNs and SNNs, with and without knowledge-distillation (KD).}
        \label{tab:resnet_perf}
    \end{minipage}
    \hfill
    \begin{minipage}[t]{\columnwidth}
        \centering
        \resizebox{\columnwidth}{!}{
        \begin{tabular}{|c|c|c|c|c|}
            \hline
            \textbf{Type} & \textbf{Architecture} & \textbf{Accuracy} & \textbf{Parameters} & \textbf{FLOPS} \\ \hline
            ANN & VGG19          & 92.22\%  & 20M  & 398M MACs \\ \hline
            SNN & 1xVGG11 (\textit{without} KD) & 87.7\% & 9.3M & 22.5M ACs \\ \hline
            SNN & 1xVGG11          & 88.21\%  & 9.3M  & 22.5M ACs \\ \hline
            SNN & 2xVGG5         & 87.54\%  & 3.6M   & 17.6M ACs \\ \hline
            SNN & 4xVGG5mini     & 87.94\%  & 3.2M  & 18.4M ACs \\ \hline
        \end{tabular}
        }
        \vspace{0.2cm}
        \caption{Validation accuracy, total parameter count and estimated FLOPs for a single forward pass of VGG-based architectures, comparing ANNs and SNNs, with and without knowledge-distillation (KD).}
        \label{tab:vgg_perf}
    \end{minipage}
\end{table*}

\begin{figure*}[!h]
    \centering
    \includegraphics[width=\linewidth]{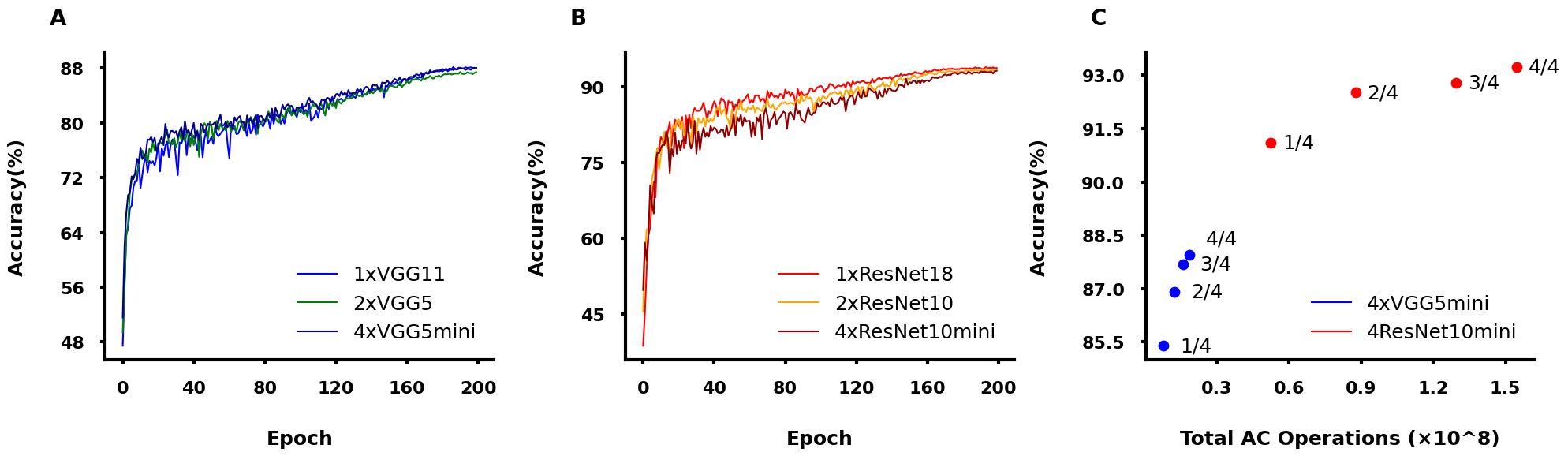}
    \caption{Performance of VGG and ResNet Architectures on CIFAR-10. Architectures of two- and four-student ensembles achieve comparable performance to the single-student SNN with only a minor reduction in accuracy. The total AC operations, considering the entire ensemble, and evaluation accuracy of the two architectures using several students active during inference. The notation ``X/Y" indicates the number of active students in an ensemble of Y students.}
\end{figure*}

\subsection{Benefits of Disentanglement}
To assess the effectiveness of the feature partitioning techniques proposed in Section \ref{sec:disentangle}, we re-evaluate the ensembles described in Section \ref{sec:num_students}. Our primary objective is to reduce the performance gap between the ANN teacher model and our Spiking Neural Ensembles. When the teacher model is frozen we employ agglomerative clustering for feature-space partitioning, assigning feature subsets to student models with different capacities.
Agglomerative Clustering outperforms both k-means clustering and fixed partitioning (see Table \ref{tab:accuracy_table}). For this, we utilize the implementation in the scikit-learn library with complete linkage \cite{b28}.

In case that students with equal capacity are required, we employ a variation of the k-means clustering. After the initial k-means clustering, we apply a greedy ``reallocation" heuristic to balance cluster sizes. Specifically, for each cluster \(c\) with a size deficit, we identify the number of elements required to achieve the target size and reassign elements from clusters of excess size. Elements to be reassigned are selected based on their proximity to \(c\)'s centroid. This way we balance the clusters, ensuring equal capacity across students. While the performance in this setup is slightly lower than the unequal-capacity case, it still demonstrates improvement over fixed feature partitioning.

\begin{table}[h]
\centering
\renewcommand{\arraystretch}{1.5}
\resizebox{\columnwidth}{!}{ 
\begin{tabular}{|c|c|c|c|c|}
\hline
\textbf{Architecture} & \textbf{Fixed} & \textbf{Agglomerative} & \textbf{K-means} & \textbf{Teacher Fine-tuning} \\ \hline
2xVGG5                & 87.54\%           & 88.21\%                                  & 88.12\%  & \textbf{89.01\%}                 \\ \hline
4xVGG5mini            & 87.94\%           & 89.11\%                                   & 88.42\% & \textbf{90.38\%}
\\ \hline
2xResNet10            & 93.03\%           & 93.18\%                                   & 93.57\% & \textbf{93.83\%}
\\ \hline
4xResNet10mini         & 93.22\%           & 93.49\%                                   & 93.32\% & \textbf{93.66\%}
\\ \hline
\end{tabular}
}
\vspace{0.2cm}
\caption{Comparison of Accuracy Across Different Architectures and Clustering Methods. Compared to SNE with fixed, k-means or agglomerative clustering, SNE with fine-tuned teacher achieves an improved accuracy.}
\label{tab:accuracy_table}
\end{table}

The teacher fine-tuning method (as illustrated in Fig. \ref{fig:teacher_ft}) achieves the best results, improving accuracy by 2.4\% in a four-student ensemble compared to fixed partitioning (see Table \ref{tab:accuracy_table}). While this approach introduces a computational overhead, requiring an additional 10 to 20 epochs of teacher training, the teacher effectively optimizes the additional loss term without compromising its classification accuracy. 
Notably, the introduced loss function \ref{eq:10} has a global optimum which the teacher successfully reaches. The teacher, through the disentanglement process, aims to position the points (normalized subvectors of each feature) as far apart as possible on the surface of a unit sphere in \(R^{D/N}\). For instance, when \(N=4\), geometrically, the optimal arrangement occurs when these 4 points form the vertices of a regular tetrahedron, maximizing pairwise distances to \(\sqrt{8/3}\), which is the global optimum of \ref{eq:10} for \(N=4\).
Similarly, when \(N=2\), the optimal arrangement occurs when the two subvectors are anti-parallel resulting in a maximum distance and global optimum value of 2.
This fine-tuning procedure disentangles the information in the teacher's features, making each assigned sub-feature more compact and easier for students to replicate. This improved feature separation likely explains the observed increase in accuracy.

\begin{table*}[!htbp]
    \centering
    \scriptsize
    \begin{tabular}{|p{2cm}|p{1.5cm}|p{1.5cm}|p{1.5cm}|p{1.5cm}|p{1.5cm}|p{1.5cm}|p{1.5cm}|p{1.6cm}|}
        \hline
        & \multicolumn{2}{c|}{\textbf{ANN}} & \multicolumn{6}{c|}{\textbf{SNN}} \\ \cline{2-9}
        \textbf{Type of Noise} & \textbf{VGG19} & \textbf{ResNet18} & \textbf{VGG11} & \textbf{2xVGG5} & \textbf{4xVGG5mini} & \textbf{ResNet18} & \textbf{2xResNet10} & \textbf{4xResNet10mini} \\ \hline
        No noise & 92.2 & 94.26 & 87.78 & 87.56 & 87.94 & 93.86 & 93.05 & 93.29  \\ \hline
        
        Gaussian ($\sigma$ = 0.01) & 91.9 $\pm$ 0.03 & 93.60 $\pm$ 0.02 & 87.64 $\pm$ 0.06 & 87.32 $\pm$ 0.03 & 87.74 $\pm$ 0.04 & 93.42 $\pm$ 0.05 & 92.74 $\pm$ 0.04 &  92.6 $\pm$ 0.03 \\ \hline
        
        Gaussian ($\sigma$ = 0.03) &  85.92 $\pm$ 0.05 & 85.77 $\pm$ 0.05 & 85.41 $\pm$ 0.05 & 83.96 $\pm$ 0.04 & 85.12 $\pm$ 0.06 & 87.62 $\pm$ 0.04 & 84.89 $\pm$ 0.09 &  85.27 $\pm$ 0.05 \\ \hline
        
        Gaussian ($\sigma$ = 0.05) & 70.89 $\pm$ 0.01 & 68.44 $\pm$ 0.08 & 73.88 $\pm$ 0.1 & 70.17 $\pm$ 0.06 & 74.13 $\pm$ 0.06 & 73.79 $\pm$ 0.08 &  69.48 $\pm$ 0.1 & 70.69 $\pm$ 0.06 \\ \hline

        Gaussian ($\sigma$ = 0.07) & 52.57 $\pm$ 0.11 & 47.14 $\pm$ 0.11 & 49.56 $\pm$ 0.11 & 48.25 $\pm$ 0.09 & 58.92 $\pm$ 0.1 & 54.22 $\pm$ 0.11 & 53.10 $\pm$ 0.06 & 55.08 $\pm$ 0.07 \\ \hline
    \end{tabular}
    \vspace{0.2cm}
    \caption{Accuracy (\%) of VGG and ResNet architectures with different amounts of Gaussian noise ($\mu=0$) added in the input dataset. Values are reported as Mean $\pm$ Standard Error of the Mean.}
    \label{tab:noise_perf}
\end{table*}

\subsection{Computation-Performance Trade-off Using Dropout}
The student ensemble offers the advantage of dynamically adjusting the number of students active depending on the need for more accuracy or less energy consumption. In order to utilize the flexibility of the ensemble we propose two methodologies.

The first methodology utilizes the \textit{complete} ensemble throughout the training process. During the evaluation phase, we employ a \textit{stochastic selection} mechanism that operates by \textit{randomly} sampling \textit{K} indexes from the range [1, \textit{N}] without replacement, where \textit{N} represents the total number of student models in the ensemble and \textit{K} the \textit{target} number of active models. These sampled indexes determine which student models will be \textit{activated for the current batch}, while the remaining \textit{N-K }models remain dormant. The input is fed only into the selected students, generating their respective output features, while zero vectors are assigned to the output feature spaces of the inactive models. This dynamic activation pattern is regenerated for \textit{each subsequent batch}, ensuring a diverse mix of active student models throughout the evaluation process. This stochastic approach provides a key advantage: by randomly selecting subsets from the full-range of possible student-model combinations, it generally produces performance levels that approximate the \textit{mean} of all possible combinations.

The second methodology utilizes the \textit{dropout} technique for both the evaluation phase and the training phase. Its key difference from the first is that for this methodology we determine \textit{before training} the number of students that will be active. Here the ensemble learns to utilize a subset of students during training. This results to an improvement of the ensemble's accuracy but at the cost of an increase in the number of AC operations of each student (see Table \ref{tab:dropout_tradeoff_train}). 
We observe that, in response to the reduced number of students, the ensemble compensates by increasing the neuronal firing rates, which in turn increases the AC operations of the ensemble.

The second methodology is advantageous when the number of active students during evaluation/testing is expected to remain relatively stable and is known in advance. 
As shown in Tables \ref{tab:dropout_tradeoff_eval} \ref{tab:dropout_tradeoff_train}, our SNEs demonstrate robustness to the number of students activated during evaluation. As the number of active students and consequently the number of operations needed decreases the accuracy decreases only slightly. This offers flexibility in the performance vs. accuracy trade-off. In the case of one active student, we observe a 65\% decrease in energy consumption with only 2.07\% degradation in accuracy. The experiments performed to examine the impact of the dropout consider a \textbf{fixed} feature-to-students partitioning.

\begin{table}[h]
    \centering
    \scriptsize 
    \renewcommand{\arraystretch}{1.3} 
    \begin{tabular}{|c|c|c|c|c|}
        \hline
        & \multicolumn{2}{c|}{\textbf{4xVGG5mini}} & \multicolumn{2}{c|}{\textbf{4xResNet10mini}} \\ \cline{2-5}
        \textbf{Active Students} & \textbf{Accuracy} & \textbf{AC Ops.} & \textbf{Accuracy} & \textbf{AC Ops.} \\ \hline
        4 & 87.94\% & 18.5M & 93.23\% & 154.8M \\ \hline
        3 & 86.57\% & 13.7M & 92.07\% & 116.2M \\ \hline
        2 & 82.67\% & 9.2M  & 90.00\% & 79.7M  \\ \hline
        1 & 75.71\% & 4.6M  & 84.27\% & 37.4M  \\ \hline
    \end{tabular}
    \vspace{0.2cm}
    \caption{Dropout tradeoff for 4xVGG5mini and 4xResNet10mini architectures. This table shows the accuracy and approximate AC operations as the number of active students decreases. All Architectures are trained with all of the students active and only evaluation utilizes a subset.}
    \label{tab:dropout_tradeoff_eval}
\end{table}

\begin{table}[h]
    \centering
    \scriptsize 
    \renewcommand{\arraystretch}{1.3} 
    \begin{tabular}{|c|c|c|c|c|}
        \hline
        & \multicolumn{2}{c|}{\textbf{4xVGG5mini}} & \multicolumn{2}{c|}{\textbf{4xResNet10mini}} \\ \cline{2-5}
        \textbf{Active Students} & \textbf{Accuracy} & \textbf{AC Ops.} & \textbf{Accuracy} & \textbf{AC Ops.} \\ \hline
        4 & 87.94\% & 18.5M & 93.23\% & 154.8M \\ \hline
        3 & 87.69\% & 15.9M & 92.79\%  &  129.5M \\ \hline
        2 & 86.91\% & 12.2M  & 92.51\% & 87.6M  \\ \hline
        1 & 85.39\% & 7.8M  & 91.16\%  & 52.6M  \\ \hline
    \end{tabular}
    \vspace{0.2cm}
    \caption{Dropout tradeoff for 4xVGG5mini and 4xResNet10mini architectures. Both training and evaluation is performed with a random subset of students active.}
    \label{tab:dropout_tradeoff_train}
\end{table}

\begin{figure}[!ht]
    \centering
    \includegraphics[width=\linewidth, height=5cm]{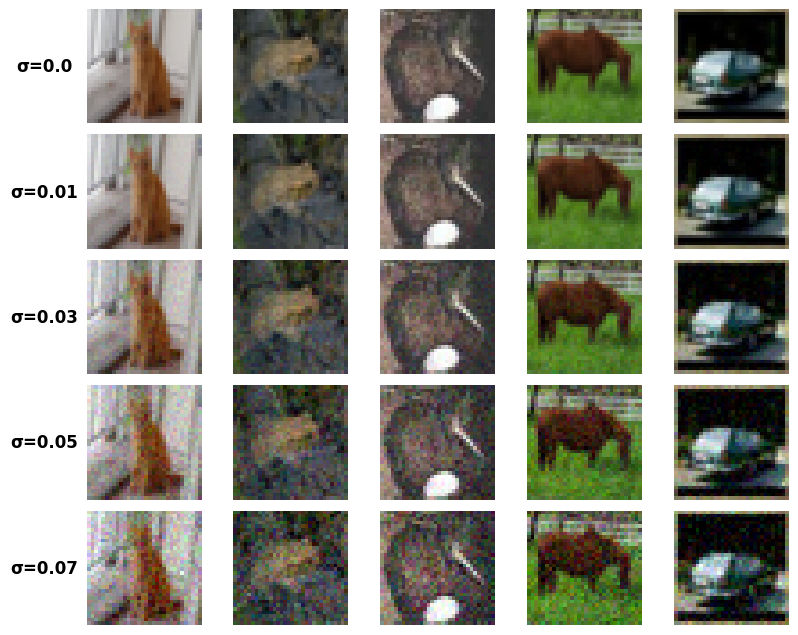}
    \caption{CIFAR-10 dataset with different amounts of Gaussian Noise ($\mu=0, \sigma$).}
    \label{fig:noisy_cifar}
\end{figure}
\subsection{Performance Under Noise}
We examine the performance of the SNE in the presence of noise added in the input dataset, during the testing phase. During training, the input dataset CIFAR-10 was without any noise. 
We tested various VGG and ResNet architectures with input under different levels of noise. Specifically, we add to each pixel of each image in the testing dataset, Gaussian noise with $\mu=0$ and $\sigma$ (see Figure \ref{fig:noisy_cifar}). We then pass the noisy images through the network to produce predictions. We ran each experiment 10 times and calculated the mean accuracy in each setting along with Standard Error of the Mean (SEM). We repeat these experiments for different levels of noise ($\sigma=0.01, 0.03, 0.05, 0.07$) Table \ref{tab:noise_perf}).

As expected, the higher the noise, the lower the accuracy, for each architecture. However, the SNE with four-student SNNs manifests a higher robustness compared to the original ANN VGG19 or ResNet18 architecture or the corresponding SNN architectures with one- or two-student ensemble. In accordance with \cite{b12}, the single-student SNN model demonstrates comparable performance to the ANN teacher under noisy data, suggesting that the single-student model retains much of the teacher's robustness.
Interestingly, there is a significant difference between the single-student case and the ensemble of 4-students, especially for high levels of noise. 
We hypothesize that the reduced number of parameters in each student in the 4-student ensemble, compared to the teacher ANN and to the single-student, could help mitigate overfitting and thereby enhance the robustness to noise.

\section{Conclusions and Future Work}\label{sec:disc}
We developed SNE, an innovative ensemble of spiking neural networks trained using knowledge distillation guided by a pre-trained artificial neural network. SNE can be an order of magnitude less computationally expensive than the teacher network (e.g. 18.4M vs. 398M FLOPS, under VGG-based models), while maintaining comparable performance on classification tasks, such as CIFAR-10 (e.g., 90.39\% vs. 92.2\%, respectively). It shows encouraging performance in addressing effectively the trade-off between computational efficiency and accuracy, by dynamically adjusting the number of active students in the ensemble, as well as improved robustness under noise. 
%
%

The informed partitioning (disentanglement) of the teacher's feature space has an impact on the knowledge distillation process. While clustering algorithms in the frozen teacher's feature space may yield modest gains, our results suggest that further exploration on the teacher fine-tuning procedure could further reduce the performance gap between ANN and SNN networks. Notably, the loss function introduced in Eq. \ref{eq:10}, despite being sample-specific, without explicitly enforcing compact sub-features \textit{across} samples, still yields superior results compared to alternative loss functions that take also into consideration the sub-feature proximity within the same cluster.

This work primarily considers SNN students that operate in parallel, concurrently processing the input image. An increased energy efficiency can be achieved by enabling students to share the first layers of their architecture and only differentiate at the top layers. Additionally, exploring hierarchical or sequential processing among students may further optimize performance.
We also plan to develop auxiliary networks to make sample-specific decisions about which students should remain active during inference.

In summary, SNE introduces a novel approach that addresses effectively the energy efficiency and accuracy tradeoff compared to traditional foundation AI models. By \textit{dynamically} activating a subset of shallow SNN models and leveraging knowledge distillation, enhanced with an informed-partitioning (disentanglement) of the teacher's feature space, SNE offers a promising solution for energy-constrained applications.
\section{Acknowledgments} We would like to thank Kostas Daniilidis and Nikos Komodakis for providing valuable feedback on the manuscript.  Also, we are grateful to Efstratios Gavves and Ioannis Smyrnakis for discussing various aspects of the knowledge distillation paradigms. This work has been partially supported by the project MIS 5154714 of National Recovery and Resilience Plan Greece 2.0 funded by the European Union under the NextGenerationEU Program, from the European Union’s Horizon 2020 research and innovation program under the Marie Skłodowska-Curie grant agreement No 101007926, and from the Hellenic Foundation Research Institute (HFRI), the neuron-AD project number 04058 and neuronXnet project number 2285 (PI: Maria Papadopouli).

\bibliographystyle{ieeetr}

\vspace{12pt}

\end{document}